\documentclass[conference]{IEEEtran}
\IEEEoverridecommandlockouts
\usepackage{cite}
\usepackage{amsmath,amssymb,amsfonts}
\usepackage{algorithmic}
\usepackage{graphicx}
\usepackage{textcomp}
\usepackage{xcolor}
\def\BibTeX{{\rm B\kern-.05em{\sc i\kern-.025em b}\kern-.08em
    T\kern-.1667em\lower.7ex\hbox{E}\kern-.125emX}}

\usepackage{booktabs}
\usepackage{dsfont}
\usepackage{cite}
\usepackage{multirow}
\usepackage{arydshln}
\usepackage{tabularray}
\usepackage{colortbl}
\newcommand{\method}{{OLE}}
\definecolor{light_cyan}{rgb}{0.88,1,1}
\begin{document}

\title{Zero-Shot Out-of-Distribution Detection with Outlier Label Exposure\\
}
\author{\IEEEauthorblockN{Choubo Ding}
\IEEEauthorblockA{\textit{Australian Institute for Machine Learning} \\
\textit{The University of Adelaide}\\
Adelaide, Australia \\
choubo.ding@adelaide.edu.au}
\and
\IEEEauthorblockN{Guansong Pang}
\IEEEauthorblockA{\textit{School of Computing and Information Systems} \\
\textit{Singapore Management University}\\
Singapore \\
gspang@smu.edu.sg}}


\maketitle

\begin{abstract}
As vision-language models like CLIP are widely applied to zero-shot tasks and gain remarkable performance on in-distribution (ID) data, detecting and rejecting out-of-distribution (OOD) inputs in the zero-shot setting have become crucial for ensuring the safety of using such models on the fly. Most existing zero-shot OOD detectors rely on ID class label-based prompts to guide CLIP in classifying ID images and rejecting OOD images. In this work we instead propose to leverage a large set of diverse auxiliary outlier class labels as pseudo OOD class text prompts to CLIP for enhancing zero-shot OOD detection, an approach we called Outlier Label Exposure (OLE). The key intuition is that ID images are expected to have lower similarity to these outlier class prompts than OOD images.  One issue is that raw class labels often include noise labels, e.g., synonyms of ID labels, rendering raw OLE-based detection ineffective. To address this issue, we introduce an outlier prototype learning module that utilizes the prompt embeddings of the outlier labels to learn a small set of pivotal outlier prototypes for an embedding similarity-based OOD scoring. Additionally, the outlier classes and their prototypes can be loosely coupled with the ID classes, leading to an inseparable decision region between them. Thus, we also introduce an outlier label generation module that synthesizes our outlier prototypes and ID class embeddings to generate in-between outlier prototypes to further calibrate the detection in OLE. Despite its simplicity, extensive experiments show that OLE substantially improves detection performance and achieves new state-of-the-art performance in large-scale OOD and hard OOD detection benchmarks. Code is available at \textcolor{blue}{https://github.com/Choubo/OLE}

\end{abstract}


\section{Introduction}\label{sec:intro}

Out-of-distribution (OOD) detection is crucial for securing the safe deployment of deep learning models \cite{amodei2016concrete, dietterich2017steps}. This is because most deep learning methods are trained under a closed data distribution \cite{krizhevsky2012imagenet} and fail to handle out-of-distribution unknown inputs in real-world scenarios. The detection and rejection of these unexpected inputs are instrumental in averting potentially catastrophic decisions, e.g., in autonomous driving or medical diagnosis systems. Traditional OOD detection methods \cite{Yang_2021_ICCV, Choi_2023_CVPR, bendale2016towards, tian2022pixel,huang2021importance,liu2023residual, zaeemzadeh2021out, huang2021mos,miao2024out} rely on learning single-modal models on training data of a target dataset, which not only requires significant computational costs but also contends with various major issues such as data privacy issues on using the training data, or overconfident prediction of OOD input caused by the closed-set training.

\begin{figure}[htbp]
    \centering
    \vspace{-0.0cm}
    \includegraphics[width=\linewidth]{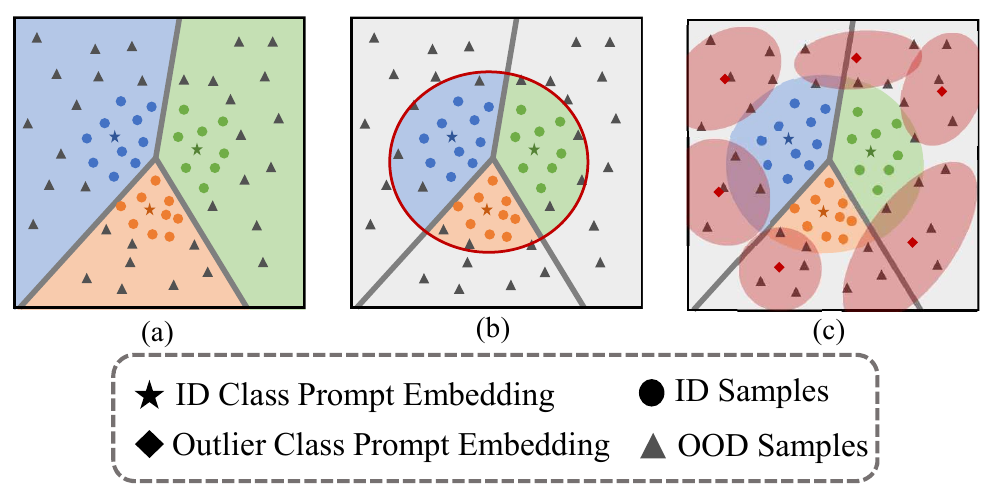}
    \vspace{-0.7cm}
    \caption{
    (a) Zero-shot classifiers do not consider OOD inputs; all OOD inputs are predicted as ID classes. (b) Current zero-shot OOD detectors help alleviate this issue but they often have overconfident predictions of 
    OOD samples due to the lack of knowledge about OOD data. (c) Our method OLE mitigates this problem by utilizing outlier class labels via text prompts to enable more OOD-informed detection. 
    }
    \label{fig:toy}
    \vspace{-0.6cm}
\end{figure}

With the remarkable generalization capabilities of large pre-trained vision-language models (VLMs) \cite{radford2021learning}, there is fast growth of attention that shifts research efforts to zero-shot tasks, in which we perform visual recognition without any tuning/training on the target data \cite{zhou2022coop,wu2023open,fang2024simple,zhu2024toward}. Zero-shot vision tasks are typically not constrained to a single closed data distribution and require multimodal information. As the VLMs like CLIP \cite{radford2021learning} are widely applied to various zero-shot vision tasks and gain remarkable performance on in-distribution (ID) data, detecting and rejecting out-of-distribution (OOD) inputs in the zero-shot setting have become crucial for ensuring the safety of using such models on the fly. 
For instance, in zero-shot classification tasks, CLIP-based zero-shot classifiers utilize a set of known ID class labels to generate text prompt embeddings and predict the image category based on the similarities between these label prompts and images in CLIP's image-text-aligned feature space. In this context, as shown in Fig. \ref{fig:toy}(a), the known label set in these classifiers represents a closed-set of classes, having the capability of predicting the images within this ID class set only, thereby lacking the capability of rejecting the OOD samples in zero-shot settings. 

Several prior works have embarked on addressing the zero-shot OOD detection problem. ZOC \cite{esmaeilpour2022zero} employs an auxiliary text generator to create potential OOD class labels for the input images. Regrettably, its efficacy is heavily contingent upon the performance of the text generator, which is not reliable when dealing with large-scale ID datasets such as ImageNet-1K \cite{deng2009imagenet}.  
Conversely, MCM \cite{ming2022delving} reevaluates the correlation between image embeddings and class label prompt embeddings in CLIP, computing OOD scores through the maximum softmax probability of the embedding similarities. Due to the lack of knowledge about OOD samples, MCM often produces high false positives, especially for the hard/near OOD images that exhibit similar features as the ID images, as shown in Fig. \ref{fig:toy}(b). To deal with this issue, very recently CLIPN \cite{wang2023clipn} augments CLIP with an additional `no' text encoder, which is trained using large auxiliary data, and achieves new state-of-the-art (SotA) performance in zero-shot OOD detection. In line with the preceding works, CLIPN attempts to detect unmodeled out-of-distribution inputs by using representations from ID class labels but fails to address the overconfidence issue.

To complement methods like CLIPN, this paper propose to leverage a large set of auxiliary outlier class labels as pseudo OOD text prompts to VLMs for providing additional knowledge about OOD representations, an approach we called \textbf{\underline{O}utlier \underline{L}abel \underline{E}xposure} (OLE). The key intuition behind OLE is that ID images are expected to exhibit lower similarity to the pseudo OOD text prompt embeddings than OOD images. Such similarity information can be utilized to enlarge the OOD scores between ID and OOD images in models like CLIPN.  

In practice, large-scale outlier class labels can be collected from the Internet in an inexpensive and efficient way. Nonetheless, due to the presence of noise labels such as the synonyms of ID labels, using the raw outlier class labels can harm the zero-shot detection performance of VLMs. To address this issue, we introduce an outlier prototype learning module to learn and refine a small set of pivotal outlier prototypes from the text prompt embeddings of these raw outlier class labels. Specifically, we first have an unsupervised modeling of the distribution of the outlier prompt embeddings via clustering and consider each resulting cluster as an outlier prototype. We then refine the outlier prototype set by eliminating the prototypes overlapping the ID class prompt embeddings to avoid the false positive detection. 

Additionally, the outlier classes and their prototypes can be loosely coupled with the ID classes, leading to an inseparable decision region between them. To tackle this problem, we also introduce an outlier label generation module that synthesizes our outlier prototypes and ID class prompt embeddings to generate and include in-between outlier prototypes to further calibrate the detection in OLE.

In summary, our main contributions are as follows:

\begin{itemize}
    \item We propose a novel approach called OLE for zero-shot OOD detection. It provides an cost-effective solution for OOD scoring in zero-shot settings by leveraging easily accessible outlier class labels to prompt VLMs for knowledge about unknown samples of a target dataset.
    \item We then propose a novel outlier prototype learning module, which compresses large-scale raw outlier class labels into a small set of pivotal outlier prototypes for efficient and accurate zero-shot detection tasks.
    \item We also introduce an outlier prototype generation module that generates OOD prototypes lying between our learned OOD prototypes and fringe ID class embeddings to further calibrate the detection in OLE.
    \item Extensive experiments on two popular benchmarks show that OLE can significantly enhance the SotA model CLIPN for both large-scale OOD detection and hard OOD detection, and achieves new SotA performance.
\end{itemize}

\section{Related Work}
\label{sec:related}
\noindent\textbf{OOD Detection.} Considerable research effort has been invested into out-of-distribution (OOD) detection for deep neural network (DNN) classifiers based on a single visual modality. One popular approach in this line involves modeling the uncertainty of pre-trained DNN classifiers and deriving OOD scores using metrics such as softmax probability~\cite{bendale2016towards, hein2019relu, liang2018enhancing}, logits~\cite{liu2020energy, sun2021react, HendrycksBMZKMS22}, feature embeddings~\cite{wang2022vim, lee2018simple, sastry2020detecting}, or gradients~\cite{huang2021importance}. 
Another popular approach is to redesign objective functions to fine-tune DNNs for both the classification and OOD detection tasks \cite{papadopoulos2021outlier, Yang_2021_ICCV, Choi_2023_CVPR, hendrycks2018deep,liu2023residual}.
In particular, unlike conventional methods that focus solely on ID samples, Outlier Exposure \cite{hendrycks2018deep} fine-tunes classifiers to learn additional OOD information by introducing a large number of outlier samples as pseudo OOD samples. These pseudo OOD samples are from either real images of auxiliary datasets \cite{ Wu_2021_ICCV, Yang_2021_ICCV, Choi_2023_CVPR,miao2024out} or synthesized outlier images \cite{lee2018training, liu2023residual,tian2022pixel,li2024learning}. Rather than using outlier images as pseudo OOD samples, our proposed OLE utilizes a plethora of readily available nouns dictionary on the Internet as pseudo OOD labels to prompt CLIP for knowledge about OOD. Moreover, the above methods often assume the availability of training ID images, which are inapplicable to zero-shot settings where we do not have such training data. 

\noindent\textbf{Zero-Shot OOD Detection.} The rise of VLMs like CLIP \cite{radford2021learning} has intensified research interest in zero-shot OOD detection. Fort et al. \cite{fort2021exploring} first apply the CLIP pre-trained model to OOD detection tasks and show CLIP's detection performance using the ground truth OOD labels. ZOC \cite{esmaeilpour2022zero} formally introduces the task of zero-shot OOD detection and learns a label generator from additional image-text datasets to generate potential OOD labels for inputs. In contrast to the above methods that rely on OOD labels, MCM \cite{ming2022delving} revisits the similarity scores of multimodal embeddings in CLIP, recalibrating similarity scores using softmax scaling to magnify the gap between ID and OOD samples. Building upon MCM,  Miyai et al. \cite{miyai2023zero} further investigate the local features of CLIP and filter out background local features irrelevant to OOD detection. Recently, CLIPN \cite{wang2023clipn} reiterates the importance of having OOD knowledge for zero-shot OOD detection, proposing to train a `no' encoder for learning prompts that have a negation semantic w.r.t. the ID labels. NegPrompt \cite{li2024negprompt} learns transferable negative prompts but it is focused on few-shot settings. Our work aligns well with previous methods \cite{esmaeilpour2022zero,wang2023clipn} in utilizing pseudo OOD information for supporting zero-shot OOD detection. Compared to these methods, our OLE is the most lightweight method, without relying on computationally-expensive extra modules like outlier label generators \cite{esmaeilpour2022zero} or additional text encoders \cite{wang2023clipn}.

\label{sec:method}
\begin{figure*}[htbp]
    \centering
    \vspace{-0.5cm}
    \includegraphics[width=0.8\textwidth]{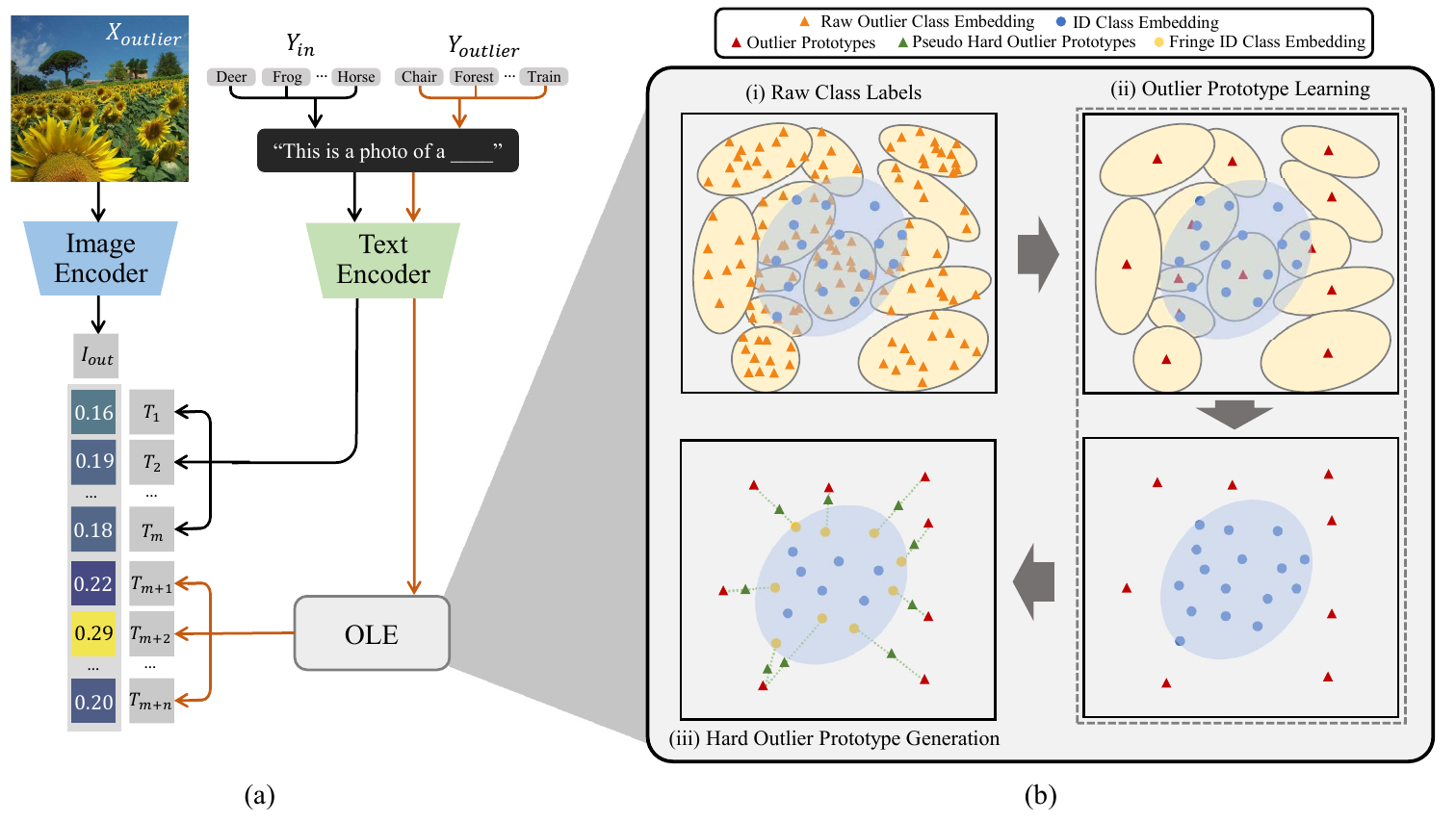}
    \vspace{-0.4cm}
    \caption{An overall framework of the proposed approach OLE. (a) present a high-level pipeline of detecting OOD inputs through the outlier label exposure in CLIP. (b) shows the process of the OPL and HOPG modules. 
    }
    \label{fig:framework}
    \vspace{-0.4cm}
\end{figure*}
\section{Proposed Approach}
\noindent\textbf{Problem Statement.} Given an in-distribution $\mathit{P}_{in}$ and an out-of-distribution $\mathit{P}_{out}$, we have a test set $\mathcal{X}=\{x_{i}\}_{i=1}^{N}$ with its samples randomly drawn from these two distributions. The ID label set $\mathcal{Y}_{in} = \{ y_{j}
\}_{j=1}^{M}$ is derived from the class semantic space of $\mathit{P}_{in}$, while the class space of $\mathit{P}_{out}$ is unknown. Let $\mathcal{I}$ and  $\mathcal{T}$ be the image encoder and the textual encoder of a pre-trained VLM,
then the goal of zero-shot OOD detection is to find a decision function $f(\mathcal{Y}_{in}, \mathcal{I},\mathcal{T}): \mathcal{X} \rightarrow \{\text{`in'},\text{`out'}\}$, in which no training ID images are available. 

Our proposed approach extends the problem by having a set of large-scale auxiliary outlier class labels $\mathcal{Y}_{\mathit{outlier}} = \{ y_{k}
\}_{k=1}^{G}$ and redesigning the decision function to $f(\mathcal{Y}_{in}, \mathcal{Y}_{\mathit{outlier}}, \mathcal{I},\mathcal{T}): \mathcal{X} \rightarrow \{\text{`in'},\text{`out'}\})$. To simulate real-world application scenarios where OOD samples are typically from unknown classes, we ensure that $\mathcal{Y}_{\mathit{outlier}}$ does not overlap with the ground truth class labels of $\mathit{P}_{out}$.

\subsection{Overview of Our Approach OLE}
We introduce the Outlier Label Exposure approach, namely OLE, that aims to leverage the superior textual understanding of the textual encoder of VLMs like CLIP for zero-shot OOD detection. The key idea is that 1) pre-trained VLMs like CLIP can yield well discriminative class embeddings when provided text prompts with the class names, and 2) in the textual-visual-aligned space of VLMs, the ID images are expected to have lower similarity to the text prompt embeddings of outlier classes, compared to the OOD images. OLE is designed to leverage those similarity information to enlarge the OOD scores of ID and OOD images. 
Although it is easy to obtain large-scale outlier class labels, it is challenging to extract relevant patterns underlying them due to the large size of the label set and the lack of supervision information about real OOD data. In OLE, we introduce two components -- outlier prototype learning (OPL) and hard outlier prototype generation (HOPG) -- to learn these patterns in an unsupervised way.

The framework of OLE is illustrated in Fig. \ref{fig:framework}(a). In the framework, a VLM first converts images and text prompts into feature embeddings through the image encoder and textual encoder. Subsequently, for ID classification, it computes the similarity between the image and textual embeddings, choosing the category with the highest corresponding value as the predicted category for the image. To detect OOD images, we introduce the OLE module, as shown in Fig. \ref{fig:framework}(b), in which the OPL and HOPG components are devised to obtain a small set of outlier prototypes with the outlier prompt embeddings as input. These outlier prototypes are then adopted to compute the similarity with the test images, which is lastly utilized to enlarge the OOD scores between test ID and OOD images.

\subsection{Outlier Prototype Learning}
The direct application of text prompts from large-scale raw outlier class labels can lead to major performance decline, wherein most ID inputs are misclassified as OOD. This issue is due to two main factors. Firstly, the large scale of outlier class labels diminishes the prediction confidence orignially attributed to the ID class labels, since the number of ID class labels is typically much less than that of outlier class labels. Secondly, these raw outlier class labels can contain considerable noise labels that have similar semantic as ID class labels.
The key challenge is how to extract relevant outlier patterns from such raw outlier labels without the knowledge of real OOD data. To this end, we introduce the outlier prototype learning (OPL) component to learn pivotal prototypes from these raw class labels, each prototype representing a concise type of outlier patterns.

The prototype learning can be attained via unsupervised clustering, whereby all outlier class label embeddings are partitioned into $K$ clusters. Each cluster's centroid is then adopted as the prototype embedding, serving as a representative for all class labels encompassed within the respective cluster. 
By using the outlier prototypes rather than the raw class labels, 
we effectively reduce the scale of the outlier prompts while minimizing the influence of noise labels, 
as shown from (i) to (ii) in Fig. \ref{fig:framework}(b).
For the clustering, we employ the Gaussian Mixture Model (GMM)\cite{reynolds2009gaussian} for fitting the distribution of these outlier prompt embeddings and extract underlying prototype patterns. GMM is a classic probabilistic model used for modeling mixture distributions and is commonly employed in unsupervised feature modeling \cite{chuang2001bayesian, ruzon2000alpha}. A GMM represents a composite distribution consisting of $K$ Gaussian sub-distributions and can be mathematically formulated as:
\vspace{-0.2cm}
\begin{equation}
    G(Q| \mu, \sigma) = \sum_{k=1}^{K} \mathcal{N}(\mathcal{T}(Q)|\mu_k, \sigma_k),
\end{equation}
where $Q$ denotes the input data that consists of the outlier class prompts $\{q_1^o, q_2^o, \cdots, q_L^o\}$, $\mathcal{T}$ is the CLIP text encoder,
$\mu_k$ and $\sigma_k$ represent the mean and variance of each sub-distribution, respectively. The $k$-th sub-distribution can be represented as:
\vspace{-0.2cm}
\begin{equation}
    \mathcal{N}(x|\mu_k, \sigma_k) = \frac{1}{\sqrt{2\pi\sigma^{2}}} e^{-\frac{(x-\mu)^{2}}{2\sigma^{2}}}.
\end{equation}

The widely used expectation-maximization (EM) algorithm \cite{moon1996expectation} is then used to estimate the parameters of GMMs \cite{wang2020weakly,liang2022gmmseg}. After that,
we use the means $\mu_k$ from all $K$ sub-distributions to obtain the outlier prototype embeddings $\mathcal{P}=\{o_1, o_2, \cdots, o_K\}$ with $o_i=\mu_i$.

Upon acquiring the prototype embeddings for outlier class labels, a new challenge arises. As depicted in the upper panel of step (ii) in Fig. \ref{fig:framework}(b), certain prototypes are closed to, or even part of, the ID class distribution, which may lead to misclassification of ID inputs.
To tackle this problem, it becomes crucial to refine these initial prototype embeddings to ensure they do not overlap with the ID class distribution. Since it is a zero-shot setting that does not provide training ID images, we propose to utilize ID class prompt embeddings to filter out those outlier prototypes. To this end, we present a straightforward and efficient solution that first calculates the similarity between outlier prototype embeddings and ID class prompt embeddings, and subsequently removes those prototype embeddings that align closely with the ID prompt embeddings using a pre-defined threshold $\lambda$. 
Specifically,
given a set of ID class prompt embedding $\mathcal{E}=\{e_j\}_{j=1}^M$ where $e_j=\mathcal{T}(q_j^{id})$ with $q_j^{id}$ being the text prompt of an ID class $j$, for each prototype embedding $o_k\in\mathcal{P}$, we define the likelihood of its overlapping to ID classes proportional to its nearest neighbor distance to the ID class prompt embeddings:

\begin{equation}
    s(o_k|\mathcal{E}) = \max_{e_j \in \mathcal{E}} (o_{k} \cdot e_{j}).
\end{equation}

We then use a filtering decision threshold $\lambda$ based on 
these nearest neighbor distances to eliminate noisy outlier prototype embeddings from $\mathcal{P}$. For instance, if we opt for the $p$-th percentile as our threshold criterion, then $\lambda$ can be computed as:
\begin{equation}
    \lambda = \text{percentile}(\{s(o_k|\mathcal{E})\}_{o_k \in \mathcal{P}}, p).
\end{equation}

The outlier prototypes with their $s(o_k|\mathcal{E})$ exceeding $\lambda$ are considered aligned closely with the ID data distribution and thus they are removed from $\mathcal{P}$, as shown in the bottom panel in step (ii) in Fig. \ref{fig:framework}(b). With this refinement process, we have successfully obtain a set of highly relevant outlier prototypes that have a minimal overlapping with the ID images while remaining effective in detecting OOD images.

\begin{figure}[tb]
    \centering
    \vspace{-0.0cm}
    \includegraphics[width=0.8\linewidth]{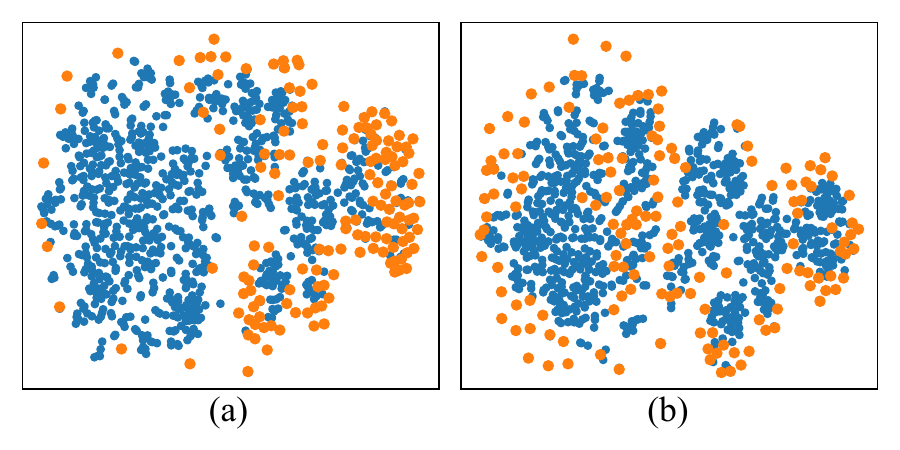}
    \vspace{-0.5cm}
    \caption{\textbf{t-SNE Visualization of the ID class prompt embeddings of ImageNet-1K.} Orange points denote the selected fringe ID classes. (a) 
    The selected fringe ID class embeddings that have the largest average distance to all ID class embeddings.  (b) The fringe ID class embeddings are identified within their clusters.}
    \label{fig:boundary}
    \vspace{-0.5cm}
\end{figure}

\subsection{Hard Outlier Prototype Generation}

One issue with the OPL component is that the resulting compressed outlier prototypes can often be loosely coupled with the ID classes, which can lead to an inseparable decision region between the ID classes and the outlier prototypes. This section introduces the HOPG module based on the idea of mixup \cite{zhang2018mixup} to address this issue. 

The mixup \cite{zhang2018mixup} method is widely used in image recognition, which generates pseudo-samples through the convex combination of two given training samples. When these two samples consist of a positive and a negative sample, mixup can generate samples that lie between the two classes, enabling the establishment of a more accurate decision boundary \cite{yang2023mixood,zhu2023openmix, zhang2023mixture}. Inspired by this, we propose a prompt embedding mix method to generate a set of pseudo outlier prototypes to calibrate the OOD detection in the aforementioned inseparable decision region. To this end, we select ID class labels that lie at the fringe of the ID data distribution and mix them with the outlier prototype embeddings to generate these psuedo outlier prototypes.

Then the first step is to identify the fringe ID class prompt embeddings. A simple approach is to calculate the distance of each ID class prompt embedding to all other 
ID class embeddings, and use the averaged farthest ID class prompt embeddings as the fringe prompt embeddings. However, this approach can introduce significant bias when applied to large-scale and unevenly distributed ID class embeddings. As shown in the Fig. \ref{fig:boundary}(left), this approach tends to select embeddings from dense regions while neglecting those from sparser regions. To rectify this issue, we propose a two-step strategy: we first cluster the ID class prompt embeddings, followed by selecting the class embeddings that are most distant from the ID class prompt embeddings within each cluster. As shown in the Fig. \ref{fig:boundary}(right), this cluster-based approach can effectively segment large-scale data into multiple uniform sub-distributions. Selecting the fringe ID class embeddings within these sub-distributions allows for a more comprehensive representation of diverse boundary regions of the ID data.

Once the fringe ID class embeddings are identified, we mix them with the compressed outlier prototypes to generate hard outlier prototypes that lie between the fringe ID class prompts and the outlier prototype prompts. Specificlly, for each fringe ID class embedding $e_j$, we select the outlier prototype that is closest to $e_j$, denoted as $o_i^j$, for the mixing:
\begin{equation}
    \tilde{o}_i = (1 - \alpha) o_i^j + \alpha e_j,
\end{equation}
where $\alpha$ is a random variable drawn from standard uniform distribution $(0, 0.5)$.
These generated hard outlier prototypes are lastly included with the compressed outlier prototypes for subsequent OOD detection.

\subsection{Inference with Outlier Label Exposure}
The resulting outlier prototype embeddings are utilized to enlarge the OOD scores of ID and OOD images during inference. 
Motivated by the superior zero-shot OOD detection performance of CLIPN \cite{wang2023clipn}, we include the learned outlier prototypes into the OOD scoring function in CLIPN to enable OOD detection.. As a derivative of the CLIP model, CLIPN trains an additional `no' text encoder to provide negative prompts supplementing the original text encoder. However, these negative prompts still rely on information derived from ID labels. Our outlier prototypes can provide complementary pesudo OOD information for CLIPN to further enhance its detection performance. Specifically, for a test image $x$, the computation of its OOD score in CLIPN is given by:

\begin{equation}\label{eqn:clipn}
    g(x) = 1 - \sum^M_{j=1}(1-p^{no}_{j}(x))p^{yes}_{j}(x),
\end{equation}
where $p_j^{yes}$ and $p_j^{no}$ represent the prediction probabilities of $x$ belonging to the $j$-th ID class yielded from the original text encoder and the `no' text encoder, respectively. To incorporating the outlier prototypes from OLE into CLIPN, we redefine the ID class prediction probabilities $p^{yes}$ from the original text encoder that takes into account of the similarity of $x$ to these outlier prototypes, in addition to the similarity to the ID class prompt embeddings. Specifically, assuming we have $G$ outlier prototypes in the end, $\{o_k\}^G_{k=1}$, the redefined $p_j^{yes}$ score of $x$ is given as follows:
\begin{equation}
    p^{yes}_{ope}(x) = \frac{e^{<x\cdot E>/\tau}}{\sum_{j=1}^{M} e^{<x\cdot e_j>/\tau} + \sum_{k=1}^{G} e^{<x\cdot o_k>/\tau}},
\end{equation}
where $\tau$ is a temperature hyperparameter. The similarities of $x$ to the outlier prototypes is added as a normalization factor into the denominator in $p^{yes}_{ope}$  to calibrate its prediction probability.  When the test image $x$ belongs to the ID data, a lower outlier similarity is expected, whose impact on the overall probability of $p^{yes}_{ope}(x)$ is limited, resulting in a similar probability value of the original $p^{yes}(x)$ in CLIPN. In contrast, when the test image belongs to the OOD, a higher outlier similarity is expected, leading to an overall decrease in $p^{yes}_{ope}(x)$. Therefore, when plugging $p^{yes}_{ope}(x)$ to replace $p^{yes}(x)$ in Eq. \ref{eqn:clipn}, we can effectively enlarge the OOD scores of ID and OOD images.

\begin{table*}[htbp]
  \centering
  \vspace{-0.5cm}
  \caption{\textbf{Results in large-scale OOD detection with ImageNet-1K as in-distribution data.} $^{\dagger}$ indicates that the results are taken from \cite{tao2023nonparametric}, and the results of $^{\ddagger}$ are taken from \cite{miyai2023locoop}. $^*$ represents results from the original paper. The remaining results are reproduced using the same architecture. The best result per dataset is \textbf{boldfaced}, with the second-best \underline{underline}. }
  \vspace{-0.3cm}
  \setlength{\tabcolsep}{3pt}
  \scalebox{1.1}{
    \begin{tabular}{p{3.3cm}cccccccccc}
    \hline
    \multirow{2}{*}{\large\textbf{Methods}} & \multicolumn{2}{c}{\textbf{iNaturalist}} & \multicolumn{2}{c}{\textbf{SUN}} & \multicolumn{2}{c}{\textbf{Texture}} & \multicolumn{2}{c}{\textbf{Places}} & \multicolumn{2}{c}{\textbf{Average}} \\
    \cmidrule(lr){2-3} \cmidrule(lr){4-5} \cmidrule(lr){6-7}  \cmidrule(lr){8-9} \cmidrule(lr){10-11}
    & \footnotesize FPR95$\downarrow$ & \footnotesize AUROC$\uparrow$ &  \footnotesize FPR95$\downarrow$ & \footnotesize AUROC$\uparrow$ &  \footnotesize FPR95$\downarrow$ & \footnotesize AUROC$\uparrow$ &  \footnotesize FPR95$\downarrow$ & \footnotesize AUROC$\uparrow$ &  \footnotesize FPR95$\downarrow$ & \footnotesize AUROC$\uparrow$ \\
    \rowcolor{lightgray}\multicolumn{11}{c}{\textit{Fine-tuned}}\\
    KNN$^{\dagger}$ \cite{sun2022out} {\scriptsize \textcolor{gray}{[ICML'22]}} & 29.17 & 94.52 & 35.62 & 92.67 & 64.35 & 85.67 & 39.61 & 91.02 & 42.19 & 90.97 
    \\
    VOS+$^{\dagger}$ \cite{du2022towards} {\scriptsize \textcolor{gray}{[ICLR'22]}} & 31.65 & 94.53 & 43.03 & 91.92 & 56.67 & 86.74 & 41.62 & 90.23 & 43.24 & 90.86 
    \\
    NPOS$^{\dagger}$ \cite{tao2023nonparametric} {\scriptsize \textcolor{gray}{[ICLR'23]}} & 16.58 & 96.19 & 43.77 & 90.44 & 46.12 & 88.80 & 45.27 & 89.44 & 37.94 & 91.22 
    \\
    \rowcolor{lightgray}\multicolumn{11}{c}{\textit{Few-shot (16-shot)}}\\
    CoOp$^{\ddagger}$ \cite{zhou2022coop} {\scriptsize \textcolor{gray}{[NIPS'23]}} & 14.60 & 96.62 & 28.48 & 92.65 & 43.13 & 88.03 & 36.49 & 89.98 & 30.68 & 91.82 
    \\
    LoCoOp$^{\ddagger}$ \cite{miyai2023locoop} {\scriptsize \textcolor{gray}{[NIPS'23]}} & 16.05 & 96.86 & 23.44 & 95.07 & 42.28 & 90.19 & \underline{32.87} & 91.98 & 27.26 & 93.53 
    \\
    \rowcolor{lightgray}\multicolumn{11}{c}{\textit{Zero-shot}}\\
    MaxLogit \cite{HendrycksBMZKMS22} {\scriptsize \textcolor{gray}{[ICML'22]}} & 60.72 & 88.19 & 44.22 & 91.31 & 52.18 & 86.02 & 56.80 & 87.14 & 53.48 & 88.16 
    \\
    Energy \cite{liu2020energy} {\scriptsize \textcolor{gray}{[NIPS'20]}} & 79.35 & 80.99 & 62.67 & 88.79 & 65.71 & 81.00 & 72.73 & 83.85 & 70.11 & 83.66 
    \\
    MCM$^{*}$ \cite{ming2022delving} {\scriptsize \textcolor{gray}{[NIPS'22]}} & 30.91 & 94.61 & 37.59 & 92.57 & 44.69 & 89.77 & 57.77 & 86.11 & 42.74 & 90.77 
    \\
    CLIPN$^{*}$ \cite{wang2023clipn} {\scriptsize \textcolor{gray}{[ICCV'23]}} & 23.94 & 95.27 & 26.17 & 93.93 & \textbf{33.45} & \underline{90.93} & 40.83 & 92.28 & 31.10 & 93.10 
    \\
    \rowcolor{light_cyan}
    CLIPN + \method\textsubscript{i21k} & \textbf{0.89} & \textbf{99.76} & \underline{24.26} & \underline{94.86} & 41.99 & 90.68 & 33.23 & \underline{92.66} & \underline{25.09} & \underline{94.49}
    \\
    \rowcolor{light_cyan}
    CLIPN + \method\textsubscript{nouns} & \underline{7.61} & \underline{98.33} & \textbf{22.44} & \textbf{94.87} & \underline{34.70} & \textbf{92.40} & \textbf{31.73} & \textbf{92.45} & \textbf{24.12} & \textbf{94.51} 
    \\
    \hline
    \end{tabular}%
  }
  \label{tab:main_result}%
  \vspace{-0.2cm}
\end{table*}%

\section{Experiments}
\label{sec:exp}
\begin{table*}[bt]
  \centering
  \vspace{-0.2cm}
  \caption{\textbf{Results in hard OOD detection.} The `A'/`B' indicates `A' as the in-distribution with `B' as the OOD dataset.}
  \vspace{-0.3cm}
  \scalebox{1.1}{
    \begin{tabular}{>{\arraybackslash}p{2cm}>{\centering\arraybackslash}p{1.8cm}>{\centering\arraybackslash}p{1.8cm}>{\centering\arraybackslash}p{1.8cm}>{\centering\arraybackslash}p{1.8cm}>{\centering\arraybackslash}p{1.8cm}>{\centering\arraybackslash}p{1.8cm}}
    \hline
    \multirow{2}{*}{\large\textbf{Methods}} & \multicolumn{2}{c}{\textbf{CIFAR10/CIFAR100}} & \multicolumn{2}{c}{\textbf{CIFAR100/CIFAR10}} & \multicolumn{2}{c}{\textbf{ImageNet-1k/ImageNet-O}} \\
    \cmidrule(lr){2-3} \cmidrule(lr){4-5} \cmidrule(lr){6-7}  
    & \footnotesize FPR95$\downarrow$ & \footnotesize AUROC$\uparrow$ &  \footnotesize FPR95$\downarrow$ & \footnotesize AUROC$\uparrow$ &  \footnotesize FPR95$\downarrow$ & \footnotesize AUROC$\uparrow$  \\
    \hline
    MaxLoigt & 35.85 & 90.44 & 57.53 & 85.81 & \textbf{52.35} & 86.07  \\
    Energy & 45.60 & 87.01 & 71.54 & 79.20 & 58.40 & 83.06   \\
    MCM & 30.34 & 94.61 & 78.36 & 83.08 & 76.75 & 77.32   \\
    CLIPN & 21.92 & 93.03 & 55.24 & 86.03 & 54.30 & 84.36   \\
    \rowcolor{light_cyan}
    CLIPN+\method & \textbf{16.94} & \textbf{95.64} & \textbf{50.67} & \textbf{88.66} & 52.55 & \textbf{86.95}  \\
    \hline
    \end{tabular}%
  }
  \label{tab:hard_ood}%
  \vspace{-0.4cm}
\end{table*}%

\subsection{Experimental Setup}
\noindent\textbf{Datasets.} 
Datasets for two popular OOD detection tasks, including large-scale OOD detection and hard OOD detection, are used to evaluate the performance of our method.
Specifcally,  
we use ImageNet-1K \cite{deng2009imagenet} data as ID dataset for large-scale OOD detection. Unlike the CIFAR benchmarks widely used in previous works \cite{fort2021exploring, esmaeilpour2022zero}, ImageNet-1K provides higher-resolution images and a wider variety of ID classes, presenting a more significant challenge. Following previous works \cite{huang2021mos,sun2021react, ming2022delving, wang2023clipn, miyai2023locoop, tao2023nonparametric}, the same datasets from iNaturalist, SUN, Texture, and Places are used as OOD test datasets.

Hard OOD detection, a.k.a., near OOD detection, primarily focuses on scenarios where there is high similarity between the OOD and ID datasets. CIFAR10 and CIFAR100 are sampled from the Tiny Images dataset \cite{torralba200880} and exhibit high similarity from each other
and they are widely considered to be near OODs relative to each other \cite{winkens2020contrastive, fort2021exploring,esmaeilpour2022zero}. Another hard OOD benchmark is ImageNet-1K vs. ImageNet-O \cite{hendrycks2021natural}. ImageNet-O is created in \cite{hendrycks2021natural} where an adversarial filtration technique is utilized to curate an ImageNet-1K-based OOD dataset that comprises OOD images misclassified with high confidence in the classifier. 

\noindent\textbf{Implementation Details.} In our main experiments, we adopt CLIPN \cite{wang2023clipn}, a variant of CLIP, as the pre-trained VLMs. CLIPN extends the original CLIP by integrating a 'no' text encoder to handle negative prompts, with the others being the same as the original CLIP model. 
We use a ViT-B-16 backbone in CLIPN.
For the class labels, we use two sets of labels for comparison: the first consists of a list of 11,508 nouns publicly available on the Internet, and the second is the ground truth labels from ImageNet-21K. Both outlier labels have been sanitized to remove any labels overlapping with the class labels in the ID and all OOD datasets to prevent data leakage. In our outlier prototype learning, we employ a Gaussian Mixture Model (GMM) and obtain $K=500$ prototypes, and then select the furthest 10\% from the ID classes as the final outlier prototypes, i.e., a set of 50 outlier prototypes. In the outlier prototype generation, we cluster the ID class labels into 5 groups, within which we select 30 fringe ID class embeddings per cluster. In total, we identified 150 fringe ID class embeddings and correspondingly generated an equal quantity of hard outlier prototypes
using a random $\alpha$ ($0.0< \alpha< 0.5$).

\noindent\textbf{Comparison Methods.} We compare the proposed method with two sets of related SotA methods: traditional post-hoc methods that are adapted to work with VLMs in zero-shot settings: Energy\cite{liu2020energy} and MaxLogit\cite{HendrycksBMZKMS22}, and SotA zero-shot methods: MCM\cite{ming2022delving} and CLIPN\cite{wang2023clipn}. We also include the recent all-shot detection methods KNN\cite{sun2022out}, VOS+\cite{du2022towards}, and NOPS\cite{tao2023nonparametric} that require fine-tuning using the ID training data, and two few-shot detection methods CoOp\cite{zhou2022coop} and LoCoOp\cite{miyai2023locoop} that are fine-tined using only few-shot ID training samples, for an in-depth analysis of the performance. To ensure a fair comparison, 
all methods are based on the CLIP-B/16 model.

\noindent\textbf{Evaluation metrics.} We use two  evaluation metrics for OOD detection: 1) \textbf{FPR95} that evaluates the false positive rate of the OOD samples when the true positive rate of the in-distribution samples is 95\%, and 2) \textbf{AUROC} that evalautes the ranking of OOD samples using the Area Under the Receiver Operating Characteristic curve. Small FPR95 or large AUROC indicates better performance.

\begin{figure*}[htb]
    \centering
    \vspace{-0.5cm}
    \includegraphics[width=0.90\textwidth]{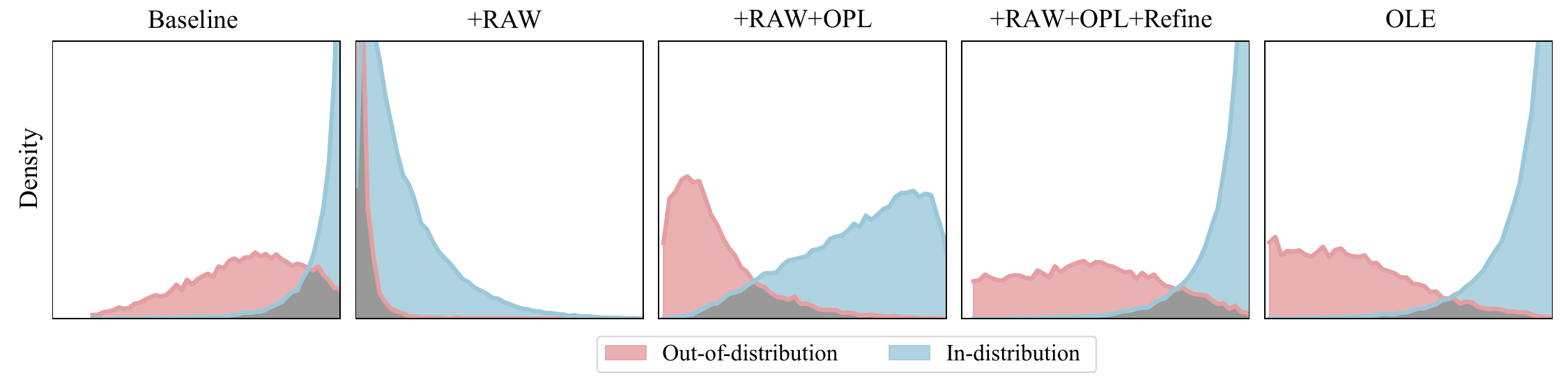}
    \vspace{-0.4cm}
    \caption{ID score density of OLE and its variants with iNaturalist as OOD data. The overlapping regions are shown in grey.
    }
    \label{fig:score_analysis}
    \vspace{-0.4cm}
\end{figure*}

\begin{table}[tb]
  \centering
  \caption{Ablation study results of OLE and its variants. The reported results are an average of four OOD datasets.}
  \vspace{-0.2cm}
  \scalebox{0.8}{
    \begin{tabular}{ccccc|cc}
    \hline
    \multicolumn{1}{c}{Baseline} & \multicolumn{1}{c}{RAW}& \multicolumn{1}{c}{OPL} & \multicolumn{1}{c}{Refine}& \multicolumn{1}{c}{HOPG}&
    \multicolumn{1}{|c}{\textbf{FPR95$\downarrow$}} &
    \multicolumn{1}{c}{\textbf{AUROC$\uparrow$}}  \\ 
    \hline
     \multicolumn{1}{c}{$\checkmark$} &\multicolumn{1}{c}{}  &\multicolumn{1}{c}{} &\multicolumn{1}{c}{} &\multicolumn{1}{c|}{} & 31.10 & 93.10
     \\
     \multicolumn{1}{c}{$\checkmark$} &\multicolumn{1}{c}{$\checkmark$} &\multicolumn{1}{c}{} &\multicolumn{1}{c}{} &\multicolumn{1}{c|}{} & 35.60 & 91.82
     \\
     \multicolumn{1}{c}{$\checkmark$} &\multicolumn{1}{c}{$\checkmark$} &\multicolumn{1}{c}{$\checkmark$} &\multicolumn{1}{c}{} &\multicolumn{1}{c|}{} & 31.45 & 92.71
     \\
     \multicolumn{1}{c}{$\checkmark$} &\multicolumn{1}{c}{$\checkmark$} &\multicolumn{1}{c}{} &\multicolumn{1}{c}{$\checkmark$} &\multicolumn{1}{c|}{} & 32.73 & 92.32
     \\
     \multicolumn{1}{c}{$\checkmark$} &\multicolumn{1}{c}{$\checkmark$} &\multicolumn{1}{c}{$\checkmark$} &\multicolumn{1}{c}{$\checkmark$} &\multicolumn{1}{c|}{} & 26.51 & 94.23
     \\
     \multicolumn{1}{c}{$\checkmark$} &\multicolumn{1}{c}{$\checkmark$} &\multicolumn{1}{c}{$\checkmark$}&\multicolumn{1}{c}{$\checkmark$} &\multicolumn{1}{c|}{$\checkmark$} & \textbf{24.12} & \textbf{94.51}
     \\
    \hline
    \end{tabular}%
    }
    \vspace{-0.4cm}
  \label{tab:ablation}%
\end{table}%

\subsection{Main Results}

\noindent\textbf{Large-scale OOD Detection.} 
The performance of OLE and other competitive methods on the ImageNet-1K benchmark with a large-scale ID class space is reported in Table \ref{tab:main_result}. Overall, the OLE significantly improves the performance of CLIPN on all metrics, reducing the FPR95 by 6.98\% and increasing the AUROC by 1.41\%. OLE achieves a new SotA performance in zero-shot detection and also outperforms fully-supervised fine-tuned and few-shot detection methods, thereby justifying the effectiveness of OPE. In the comparison of two different outlier label sets, OPE\textsubscript{nouns}, and OPE\textsubscript{i21k} have similar overall performance, with a marginal FPR95 difference of less than 1\% and a minimal AUROC discrepancy of merely 0.02\%, illustrating that OPE's effectiveness does not hinge on a specific outlier label set. The results on individual OOD datasets reveal that OPE\textsubscript{i21k} performs better on iNaturalist, while OPE\textsubscript{nouns} excels on the remaining three datasets. These variations underscore the importance of selecting appropriate outlier labels for task-specific applications. Compared to the other methods, OLE exhibits a pronounced advantage on the iNaturalist, SUN, and Places datasets.

\noindent\textbf{Hard OOD Detection.} 
Table \ref{tab:hard_ood} shows the results of three hard OOD detection benchmarks that focus on detecting OOD samples with some semantically similar features to ID samples. The results demonstrate the superior performance of OLE over the competitive methods. For example, compared to the original CLIPN, OLE reduces the FPR95 by an average of 3.76\% and increases the AUROC by an average of 2.61\%. For the two tasks where CIFAR10 and CIFAR100 are reciprocally treated as OOD data, OLE exhibits SotA 
performance, proving its effectiveness against near OOD. Furthermore, OLE exhibits significant advantages on the challenging ImageNet-O task, achieving the best AUROC and an FPR95 only second to MaxLogit.

\begin{figure}[htbp]
    \centering
    \vspace{-0.0cm}
    \includegraphics[width=\linewidth]{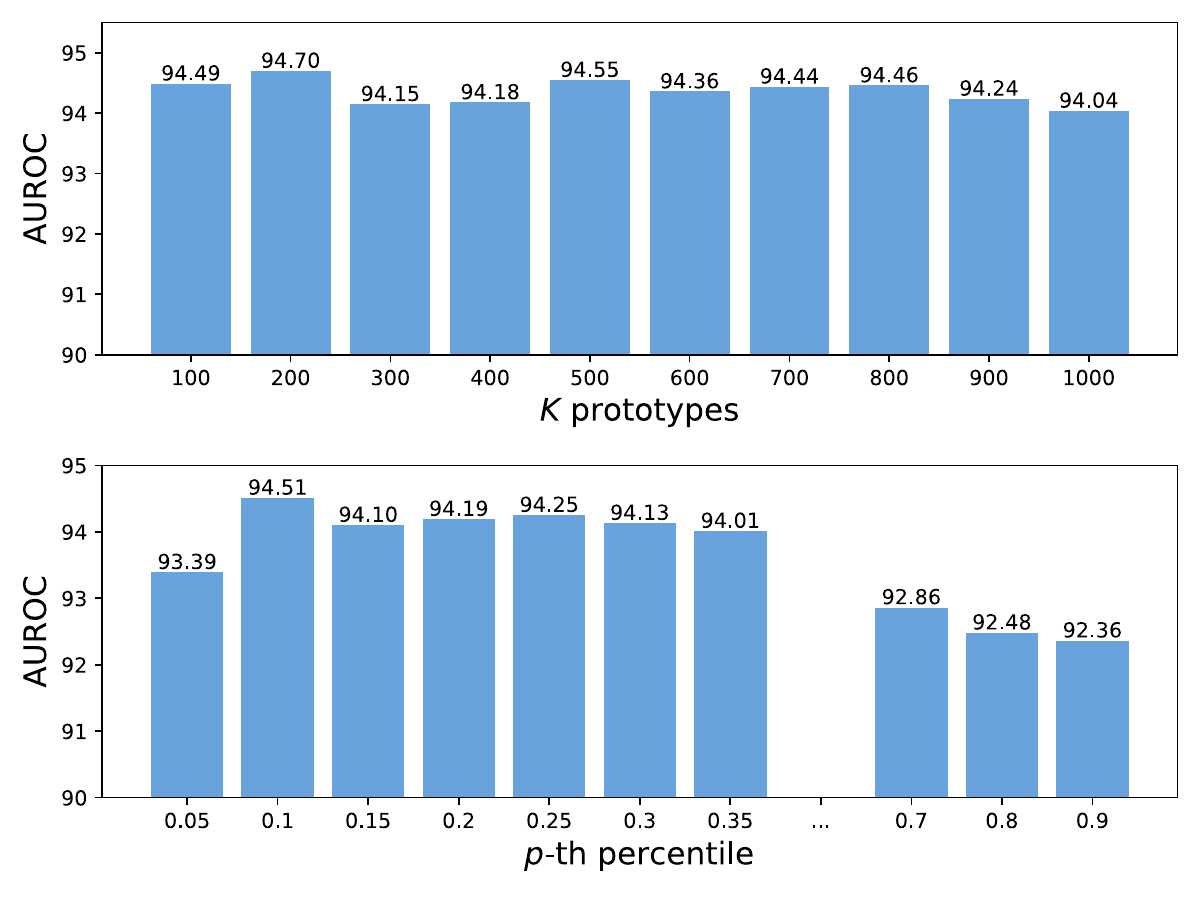}
    \vspace{-0.8cm}
    \caption{Sensitivity analysis w.r.t. (Top) the number of outlier prototypes and (Bottom) the $p$-th percentile in our OPL module.}
    \label{fig:cluster_analysis}
    \vspace{-0.4cm}
\end{figure}

\subsection{Ablation Study}

We report the results of the ablation study for OLE in Table \ref{tab:ablation}. The baseline corresponds to the original CLIPN model. 
RAW denotes the use of raw outlier class labels for OOD detection. In OLE, the OPL module learns a small set of pivotal prototypes of the outlier class labels (denoted as OPL in the table), and  we use `Refine' in Table \ref{tab:ablation} to refer to the refinement process that removes the outlier prototypes closely aligning with the ID data. 
As previously mentioned, directly utilizing raw class labels leads significant decrease in the performance, as shown by the results of RAW. This is attributed to numerous noisy labels within the raw labels. Compared to the raw class labels, the OPL and prototype refinement reduce FPR95 by 9.09\% and increase AUROC by 2.41\%. It is worth noting that applying OPL or the refinement alone yields better performance than RAW, but both are inferior to the baseline. This indicates that both methods are necessary for extracting the pivotal outlier prototypes. Furthermore, our HOPG module can effectively simulate unknown OOD classes, further reducing FPR95 by 2.39\%.

To highlight the individual contributions of the modules we propose, we analyze the change in ID score distribution for every module, the results of which are depicted in Fig. \ref{fig:score_analysis}. The baseline assigns exceedingly high ID scores to OOD samples, resulting in insufficient differentiation. After introducing the raw outlier class labels, the ID scores of all samples, including ID samples, are significantly reduced. This observation substantiates our premise that the noise labels can lead to a misclassification problem. The issue is successfully mitigated by learning the outlier prototypes OPL from the raw outlier class labels, but compared to the baseline, the distribution of ID samples is still biased to the left. This could be attributed to the presence of prototype embeddings aligned well to the ID labels. The refinement process in the outlier prototype learning not only sustains a similar ID sample distribution but also effectually diminishes the ID scores of OOD samples, thereby establishing a sharp distinction. Finally, adding the pseudo-hard outlier prototypes generated by the HOPG module further enlarges the difference between the two score distributions.

\subsection{Sensitivity Analysis}
\noindent\textbf{Sensitivity w.r.t.\  the Number of Prototypes} 
This section evaluates the sensitivity of our method OLE to different numbers of outlier prototypes and report the AUC results in Fig. \ref{fig:cluster_analysis}(Top). The results indicate that OLE can achieve stable performance with varying number of the outlier prototypes. This is mainly because the percentile-based prototype filtering in the OPL module is based on a disparate quantity setting, which helps to achieve that the retained
prototypes keep a consistent distance from the ID distribution. 
The default setting for OLE is $K = 500$ in all our experiments.

\noindent\textbf{Sensitivity w.r.t.\  the Percentile} We employ a pre-defined threshold $\lambda$ to refine the learned prototypes in the outlier prototypes learning module. The thresholds $\lambda$ are modulated by the percentile $p$, which is designed to select similar thresholds across diverse settings of prototype quantities. The sensitivity of our method to the percentile is reported in Fig. \ref{fig:cluster_analysis}(Bottom). The results confirm that retaining prototypes close to the ID classes does compromise the performance of OOD detection. On the other hand, a lower percentile, which could result in too few outlier prototypes, can also degrade the performance toward using the baseline without OLE. Notably,  the performance of OLE shows relative stability within the percentile range of $p \in [0.1, 0.35]$. This observation implies that our proposed method exhibits robustness to percentile variations within a reasonable range.

\section{Conclusions}
\label{sec:conclusion} 
In this work, we explore the idea of using a large-scale auxiliary outlier class labels in text prompts to CLIP for zero-shot OOD detection. Utilizing these auxiliary diverse class labels, we propose a novel approach for this task, called OLE, based on outlier prototype learning and generation. 
Despite the simplicity of OLE,
extensive experimental results on large-scale and hard OOD benchmarks have shown its effectiveness. This work contributes a new perspective on the design of zero-shot OOD detection algorithms. We hope this work can inspire more research on class label exposure for OOD detection to achieve more promising OOD detection models for real-world applications.





{
\small
\bibliographystyle{IEEEtran}
\bibliography{bib/IEEEfull}
}

\vspace{12pt}
\color{red}

\end{document}